%% file: bare_jrnl.tex
\documentclass[journal]{IEEEtran}

\usepackage{times}
\usepackage{latexsym}
\usepackage{CJKutf8}
\usepackage{amsmath}
\usepackage{tabularx}
\usepackage{amssymb}
\usepackage{algcompatible}
\usepackage{algorithm} 
\usepackage{booktabs}
\usepackage{multirow}
\usepackage{enumitem}
\usepackage{url}
\usepackage{color}
\renewcommand{\vec}[1]{\boldsymbol{#1}}

\usepackage{cite}

\ifCLASSINFOpdf
  \usepackage[pdftex]{graphicx}
\else
\fi
\begin{document}
%
\title{Out-of-domain Detection for Natural Language Understanding in Dialog Systems}
%
%
%

\author{Yinhe~Zheng,
        Guanyi~Chen,
        Minlie~Huang
\thanks{Yinhe Zheng is with Samsung Research China - Beijing (SRC-B), Beijing 100102, China, and Tsinghua University, Beijing 100084, China. This work was done when he works as a post-doctor in a joint program of Tsinghua University and SRC-B (e-mail: yh.zheng@samsung.com, zhengyinhe1@163.com).}
\thanks{Guanyi Chen is with the Department of Information and Computing Sciences, Utrecht University, 3584 CC Utrecht, Netherlands (email: g.chen@uu.nl).}
\thanks{Minlie Huang is with the Institute for Artificial Intelligence, State Key Lab of Intelligent Technology and Systems, Beijing National Research Center for Information Science and Technology, Department of Computer Science and Technology, Tsinghua University, Beijing 100084, China (email: aihuang@tsinghua.edu.cn).}
\thanks{This work was supported by the National Science Foundation of China (Grant No. 61936010/61876096) and the National Key R\&D Program of China (Grant No. 2018YFC0830200).}
\thanks{Manuscript received in 27-Oct-2019; revised in 33-Mar-2020.}
\thanks{\emph{Corresponding author: Minlie Huang.}}}

%
%

\markboth{IEEE/ACM Transactions on Audio, Speech, and Language Processing}
{}
%



\maketitle

\begin{abstract}
Natural Language Understanding (NLU) is a vital component of dialogue systems, and its ability to detect Out-of-Domain (OOD) inputs is critical in practical applications, since the acceptance of the OOD input that is unsupported by the current system may lead to catastrophic failure. However, most existing OOD detection methods rely heavily on manually labeled OOD samples and cannot take full advantage of unlabeled data. This limits the feasibility of these models in practical applications. 

In this paper, we propose a novel model to generate high-quality pseudo OOD samples that are akin to IN-Domain (IND) input utterances and thereby improves the performance of OOD detection. To this end, an autoencoder is trained to map an input utterance into a latent code. Moreover, the codes of IND and OOD samples are trained to be indistinguishable by utilizing a generative adversarial network. To provide more supervision signals, an auxiliary classifier is introduced to regularize the generated OOD samples to have indistinguishable intent labels. Experiments show that these pseudo OOD samples generated by our model can be used to effectively improve OOD detection in NLU. Besides, we also demonstrate that the effectiveness of these pseudo OOD data can be further improved by efficiently utilizing unlabeled data.
\end{abstract}


%
\IEEEpeerreviewmaketitle

%
%
%
%

\input{sections/introduction.tex}
\input{sections/related.tex}
\input{sections/model.tex}
\input{sections/experiment.tex}
\input{sections/conclusion.tex}

\section*{Acknowledgment}
This work was supported by the National Science Foundation of China key project with grant No. 61936010 and regular project with grand No. 61876096, and the National Key R\&D Program of China (Grant No. 2018YFC0830200).

The first author would also like to thank Wonkwang Shin, Seonghan Ryu and Kyoung-Gu Woo for fruitful discussions, and thank the support provided by the Bixby development team.

\ifCLASSOPTIONcaptionsoff
  \newpage
\fi



\bibliographystyle{bibtex/IEEEtran}
\bibliography{ref}
%

%

\begin{IEEEbiography}[{\includegraphics[width=1in,height=1.25in,clip,keepaspectratio]{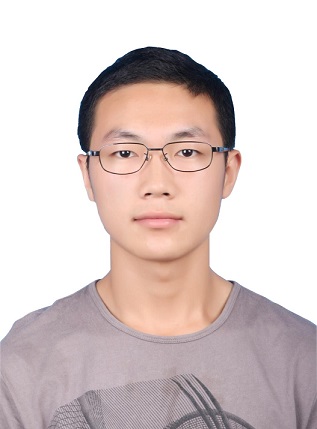}}]{Yinhe Zheng}
received his Ph.D. degree from China University of Geosciences (Beijing), Beijing, China, in 2017. He is currently working as a Post Doctor in a joint program of the Department of Computer Science and Technology, Tsinghua University, and Samsung Research China - Beijing (SRCB). His research interests include natural language processing and dialogue system, especially the tasks related to natural language understanding and natural language generation. He is the recipient of the Wuwenjun AI Award in 2019.
\end{IEEEbiography}

\begin{IEEEbiography}[{\includegraphics[width=1in,height=1.25in,clip,keepaspectratio]{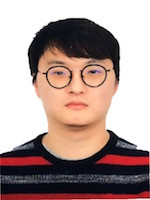}}]{Guanyi Chen}
received his M.S. degree in artificial intelligence from University of Edinburgh, Edinburgh, Scotland, in 2016. He is currently a Ph.D candidate with the Department of Information and Computing Sciences, Utrecht University, Utrecht, Netherlands. His research interests mainly lie in natural language generation, especially the computational modeling of human language production.
\end{IEEEbiography}


\begin{IEEEbiography}[{\includegraphics[width=1in,height=1.25in,clip,keepaspectratio]{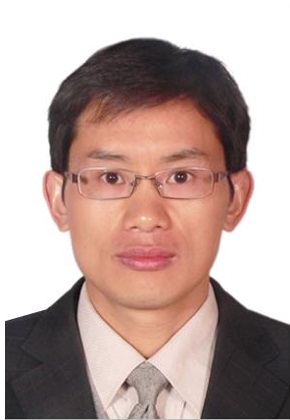}}]{Minlie Huang}
received his Ph.D. degree from Tsinghua University, Beijing, China, in 2006. He is currently an Associate Professor with the Department of Computer Science and Technology, Tsinghua University. His research interests include natural language processing, particularly in dialog systems, reading comprehension, and sentiment analysis. He has published more than 60 papers in premier conferences and journals (ACL, EMNLP, AAAI, IJCAI, WWW, SIGIR, etc.). His work on emotional chatting machines was reported by MIT Technology Review, the Guardian, Nvidia, and many other mass media. He serves as standing reviewer for TACL, area chairs for ACL 2020/2016, EMNLP 2019/2014/2011, and Senior PC members for AAAI 2017-2020 and IJCAI 2017-2020, and reviewers for TASLP, TKDE, TOIS, TPAMI, etc. He is a nominee of ACL 2019 best demo papers, the recipient of IJCAI 2018 distinguished paper award, CCL 2018 best demo award, NLPCC 2015 best paper award, Hanvon Youngth Innovation Award in 2018, and Wuwenjun AI Award in 2019. He was supported by a NSFC key project , several NSFC regular projects, and many IT companies.
\end{IEEEbiography}




\end{document}

%% file: sections/introduction.tex
\section{Introduction}\label{sec:introduction}

\IEEEPARstart{N}{atural} Language Understanding (NLU) in dialog systems such as task-oriented dialog systems and intelligent personal assistants is vital for understanding users' input to make effective human-machine interaction. An NLU module maps unstructured text inputs to structured dialog acts and has a crucial influence on the downstream processing pipelines of a dialog system. Therefore, the reliability of NLU becomes a precursor to the success of dialog systems. Recently, various deep neural network-based NLU models are proposed, and some of these models have been applied in real-world applications ~\cite{kim2018efficient,sarikaya2017technology,yoo2018data}.

Most existing neural NLU modules are built by following a \emph{closed-world} assumption \cite{Fei2016Breaking,scheirer2013toward}, i.e., the data used in training and testing phrase are drawn from the same distribution. However, such an assumption is commonly violated in practical systems that are deployed in a dynamic or open environment. Specifically, practical NLU modules often encounter \underline{o}ut-\underline{o}f-\underline{d}omain (OOD) inputs that are not supported by the system and thus not observed in the training data. Wrongly accepting these inputs and executing undesired commands may trigger catastrophic failures, particularly in risk-sensitive applications where safety is the top priority, such as robots or self-driving cars. In order to address this issue, a more realistic assumption of \emph{open-world}~\cite{scheirer2013toward,Fei2016Breaking} has been proposed. An NLU system built under this assumption should be able to not only correctly analyze \underline{i}\underline{n}-\underline{d}omain (IND) inputs but also reliably reject OOD inputs that are not supported by the system.

Various methods have been proposed to improve the OOD detection performance of neural NLU models~\cite{bendale2016towards,Ryu2018Neural,nalisnick2018do,ren2019likelihood,shafaei2018does}, and most of them follow a threshold-based protocol. Specifically, a detection score is computed for each input, and then a threshold is selected using a validation set. The inputs whose scores are lower than the threshold are considered to be OOD inputs and then rejected. A simple yet efficient approach is to use the maximum value of the Softmax output as the detection score~\cite{hendrycks2016baseline}, which has been demonstrated to work surprisingly well on the image classification tasks~\cite{liang2017enhancing}, and thus has been applied in many state-of-the-art systems~\cite{liang2017enhancing,Vyas2018out}. Further developments in this direction propose to add an extra entropy regularization (ER) term in the training objective, and significant performance improvement for OOD detection is reported when this ER term is optimized using a set of OOD data~\cite{lee2017training,hendrycks2018deep}.

However, collecting large-scale OOD data is usually difficult and expensive in practice, especially when dealing with the ever-changing \emph{open-world} environment. This limits the feasibility of the ER technique in practical applications. To address this issue, some studies~\cite{lee2017training,ryu2018out} have proposed to generate pseudo OOD samples with a generative adversarial network (GAN)~\cite{Goodfellow2014Generative}. These generated samples can be used to optimize the ER term and thus improve the OOD detection performance. However, existing approaches only work in continuous space (such as generating OOD images or continuous feature vectors), whereas the input to NLU modules is usually a sequence of discrete tokens. It is yet to be explored to generate pseudo OOD samples in discrete spaces (like natural language) that can be effectively utilized to improve the OOD detection performance. Moreover, unlabeled data (i.e., a mixture of IND and OOD samples) are usually easier to obtain in practical applications (e.g., through user logs), but rarely utilized in existing OOD detection methods. It is attractive to take advantage of these unlabeled data to improve the OOD detection performance since we can expect to gain some prior knowledge about the testing OOD distribution through these unlabeled data.

In this paper, we study how generated pseudo OOD samples and unlabeled data can facilitate OOD detection in NLU systems. We follow the simple and efficient approach to add an ER term in the training objective. However, we focus more on the way of generating high-quality pseudo OOD samples to effectively optimize this term. To this end, we propose a novel \underline{p}seudo \underline{O}OD sample \underline{g}eneration model (POG) (depicted in Figure \ref{fig:model_arch}).

The proposed model POG consists of three components: 1) a reconstruction module, 2) an adversarial generation module, and 3) an auxiliary classifier. In the reconstruction module, an encoder maps a text input into a latent code, and a decoder reconstructs the text from the latent code. A generator is then trained to produce fake latent codes, and a discriminator is trained to distinguish these fake codes from the real ones with an adversarial training process. In order to provide more supervision signals, an auxiliary classifier is further introduced to predict the correct labels associated with the reconstructed samples and to regularize the generated OOD samples to have indistinguishable labels. Experiments show that the OOD samples generated using our model can effectively improve the performance of OOD detection. We also demonstrate that unlabeled data can be used to train the reconstruction module and thus boost the effectiveness of the generated pseudo OOD samples.

To summarize, our contributions are in three folds:
\begin{enumerate}
  \item We propose a novel model to generate pseudo OOD samples. The model consists of an autoencoder, an adversarial training component, and an auxiliary classifier. 
  The generated samples can be used to effectively improve the OOD detection performance of NLU by optimizing the entropy regularization (ER) term in the training stage. 
  
  \item The proposed model can take advantage of unlabeled data to improve the effectiveness of the generated OOD samples. This makes our model more suitable for practical applications.
  
  \item We evaluate the model on two datasets. Results show that our model can significantly outperform other competitive baselines for OOD detection.
\end{enumerate}

%% file: sections/related.tex
\section{Related Work}

The problem of OOD detection has been investigated in many contexts with different alias, such as ``anomaly detection''~\cite{hendrycks2018deep}, ``one-class classification''~\cite{khan2009survey,khan2014one}, ``open-set recognition''~\cite{geng2018recent}, or ``novelty detection''~\cite{kliger2018novelty}. Significant results have been achieved by conventional methods in low-dimensional spaces~\cite{Pimentel2014review,khan2014one}, and some of these methods have also been applied to NLU systems~\cite{Lane2006Out,tur2014detecting}.

Some recent neural models use only IND data for OOD detection. Most of these methods follow the threshold-based protocol, and various approaches for calculating the detection scores are devised. Popular approaches include modeling the probability density~\cite{nalisnick2018do,pidhorskyi2018generative}, computing reconstruction losses~\cite{an2015variational,golan2018deep,Ryu2018Neural}, using classifier ensembles~\cite{Vyas2018out,shu2017doc}, applying Bayesian models~\cite{malinin2018predictive}, relying on distances to nearest-neighbors~\cite{ren2019likelihood,gu2019statistical}, or even explicitly learning a detection score~\cite{devries2018learning}. Some KNN-based methods are also applied to handle text inputs~\cite{Oh2018Out,xu2019open}. However, most of these methods are computationally expensive either in training or inference, and cannot take full advantage of unlabeled data to improve the OOD detection performance. Some of these methods also require a tremendous amount of memories as the number of classes increases~\cite{xu2019open}. All these disadvantages limit the feasibility of these methods in practical applications.

Another type of neural-based OOD detection model aims to utilize a set of OOD data in the training phase. Specifically, a special ``OOD'' label is added in a binary or multi-class classifier (e.g.,~\cite{kim2018joint}), and the inputs that fall into this special ``OOD'' class is rejected. However, the feasibility of this naive approach is limited in practice since suitable OOD data are usually hard to collect, and incorporating too many irrelevant OOD samples in training may cause a serious issue of data imbalance. Further, the OOD distribution is usually too broad to capture with these limited OOD data. 

Our study is also related to a large number of works on controllable text generation~\cite{bowman2015generating,bahuleyan2019stochastic,kim2017adversarially}, some of which also involve an adversarial training process~\cite{subramanian2018towards,hu2017toward} and controls the generated content. However, most previous studies for controllable text generation aim to model a smooth representation space and produce fluent utterances within the data distribution, whereas our model targets at improving the OOD detection performance of an NLU system and tries to generate effective OOD samples that are not in the given data distribution.

Another branch of related studies is the \underline{P}ositive and \underline{U}nlabeled (PU) learning~\cite{bekker2018learning}, in which a learning algorithm has only access to positive examples and unlabeled data. The difference between our study and PU learning models is that we aim to reject all samples that are not from IND classes. There is no guarantee that the unlabeled data can cover the entire OOD distribution, which is usually too large to tackle. However, the negative distribution considered in PU learning models is assumed to be completely covered by the unlabeled data.

There are two closely relevant studies from Ryu \emph{et al}.~\cite{ryu2018out} and Lee \emph{et al}.~\cite{lee2017training}. These studies utilize a GAN based generator to produce OOD samples when building the OOD detector. The major difference between our study and these works is that we focus on generating discrete token sequences, whereas previous works can only generate samples in the continuous space (e.g., images or continuous feature vectors). Moreover, our model can also utilize unlabeled data to improve OOD detection performance, while previous works only use IND data.

%% file: sections/model.tex
\section{Model}
\subsection{Task Definition}
In this study, we aim at improving the OOD detection performance of a practical NLU module in a dialogue system. We focus on the intent classification task since it is the most important role of an NLU module. These OOD inputs that are not supported by the current system are rejected once they are detected in the intent classification process.

Our task can be formally defined as below: Given a set of IND data, which are drawn independently from an IND distribution $\mathcal{P}_{ind}$, i.e,
$$\mathcal{D}_{ind} = \{(x_1, y_1), \cdots, (x_n, y_n)\} \sim \mathcal{P}_{ind},$$
and a set of unlabeled data, which are either drawn from the IND distribution $\mathcal{P}_{ind}$ or the OOD distribution $\mathcal{P}_{ood}$, i.e.,
$$\mathcal{D}_{mix} = \{\hat{x}_1, \cdots, \hat{x}_n\} \sim \mathcal{P}_{ind} \textrm{ or } \mathcal{P}_{ood},$$
where $x_i$ and $\hat{x}_i$ represent utterances, and $y_i \in \{l_1, \cdots, l_m\}$ is $x_i$'s label (i.e., intent type). We aim to build an intent classifier which can (1) reject the input $x$ if $x$ is drawn from $\mathcal{P}_{ood}$, and (2) predict the correct intent type of $x$ if it is from $\mathcal{P}_{ind}$.

Note that in most cases, $\mathcal{P}_{ood}$ is not known, or its underlying space is too large to explore. We cannot expect to capture the entire $\mathcal{P}_{ood}$ with the limited OOD data sampled from $\mathcal{D}_{mix}$. However, we can expect to gain some prior knowledge about the OOD distribution with the help of $\mathcal{D}_{mix}$ to improve the performance of OOD detection.

\subsection{Classifier with Entropy Regularization}\label{sec:cls_ent}

In this study, the threshold based approach is used for detecting OOD inputs. Similar to the method introduced in~\cite{hendrycks2016baseline}, we use the maximum value of the Softmax output as the OOD detection score. Specifically, the intent classifier is built with a Softmax output layer to predict an $m$-dimension distribution $P_\theta(\vec{y}|x)$ for each input utterance $x$:
$$P_\theta(\vec{y}|x) = [P_\theta(y=l_1|x), P_\theta(y=l_2|x), \cdots, P_\theta(y=l_m|x)]$$
The detection score for $x$ is obtained by:
\begin{equation}\label{eq:ood_score}
    Score(x) = \max_{i\in \{1,2,...,m\} } P_{\theta}(y=l_i|x)
\end{equation}
where $\theta$ denotes the parameters of the classifier, and $m$ is the number of intent types in the NLU module. 

In order to determine whether an input utterance $x$ is from $\mathcal{P}_{ind}$ or $\mathcal{P}_{ood}$, a threshold $t$ is chosen (usually based on the validation set). The input $x$ is regarded as an IND sample and further processed by the system if $Score(x) \geq t$, otherwise it is determined to be an OOD sample and thus rejected by the system. Therefore, it is desirable for IND inputs to obtain higher detection scores, while OOD inputs to obtain lower detection scores.

Usually, the intent classifier is trained by minimizing the cross-entropy loss:
\begin{equation}
    \mathcal{L}_{ce}(\theta) = \mathop{\mathbb{E}}_{(x_i,y_i) \sim \mathcal{P}_{ind}}[-log P_{\theta}(y=y_i|x_i)]
\end{equation}
Minimizing $\mathcal{L}_{ce}(\theta)$ on $\mathcal{D}_{ind}$ enforces the classifier to produce confident predictions on IND samples, which leads to high detection score. However, neural models trained based on the cross-entropy loss tend to be overconfident~\cite{guo2017calibration}, which results in the fact that samples from $\mathcal{P}_{ood}$ may also receive a high detection score~\cite{liang2017enhancing} if they share similar patterns and phrases with some IND samples. For example, consider a smartphone without a Bluetooth module. The intelligent assistant equipped on the phone may receive the OOD utterance ``Turn on the Bluetooth please'' from the user, who may not be familiar with the precise capability of the smartphone. This OOD input may receive a high detection score since the pattern ``Turn on ... please'' is commonly used in other IND inputs. This makes the IND and OOD inputs indistinguishable with the detection scores. In order to address this issue, a regularization term can be added to enforce a high entropy for the samples from $\mathcal{P}_{ood}$, i.e.,
\begin{equation}\label{eq:entropy}
    \mathcal{L}_{ent}(\theta) = \mathop{\mathbb{E}}_{\hat{x} \sim \mathcal{P}_{ood}}[- \mathcal{H}(P_{\theta}(\vec{y}|\hat{x}))]
\end{equation}
where $\mathcal{H}$ is the Shannon entropy of the predicted distribution. This term is similar to the \emph{confidence loss} used in~\cite{lee2017training} as it enforces the predicted distribution of OOD inputs closer to the uniform distribution, and thus leads to lower detection scores for OOD inputs.

The total loss for the intent classifier is
\begin{equation}\label{eq:loss}
    \mathcal{L}_{cls}(\theta) = \mathcal{L}_{ce}(\theta) + \alpha\mathcal{L}_{ent}(\theta)
\end{equation}
where $\alpha$ is a hyper-parameter to balance the contribution of the entropy regularization term. In this study, we set $\alpha=1$\footnote{We also tried other values for $\alpha$, but the experiments show that the final result is not sensitive to this value.}.

Note that in the original work of ~\cite{hendrycks2018deep}, $\mathcal{L}_{ent}(\theta)$ is optimized with samples drawn from $\mathcal{P}_{ood}$. Ideally, we should sample all types of OOD inputs if we cannot obtain any prior knowledge for $\mathcal{P}_{ood}$. However, this is often infeasible, if not impossible, in practical applications. We thus propose to tackle this issue with a pseudo OOD sample generation module, which will be detailed in the next section.

\begin{figure*}[!t]
  \centering
  \begin{tabular}{c}
    \includegraphics[width=14cm]{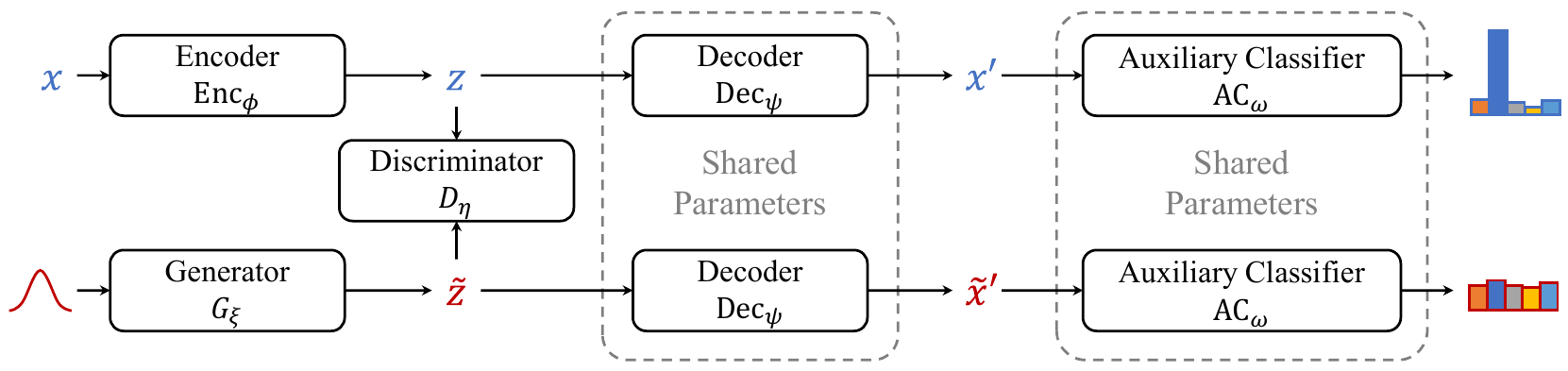} 
  \end{tabular}
  \caption{Overall architecture of the pseudo OOD sample generation (POG) model. An encoder Enc$_\phi$ transforms an input utterance $x$ to a latent code $z$. A decoder Dec$_\psi$ reconstructs $x'$ from $z$. A generator $G_\xi$ is built to map a Gaussian noise to a fake latent code $\tilde{z}$ and a discriminator $D_\eta$ distinguishes the fake code $\tilde{z}$ and the real code $z$ with an adversarial training process. An auxiliary classifier AC$_\omega$ is trained to predict the correct label associated with $x'$, and $G_\xi$ is regularized by the gradients derived from  AC$_\omega$ which enforces utterance $\tilde{x}'$ generated from $\tilde{z}$ to follow an uniform distribution.}
  \label{fig:model_arch}
\end{figure*}

\subsection{Pseudo OOD Sample Generation}

In this section, we present a novel pseudo OOD sample generation (POG) model, which employs an adversarial generation process. The produced pseudo OOD samples can be used to evaluate the entropy regularization term $\mathcal{L}_{ent}(\theta)$, thereby improving the performance of OOD detection. 

\subsubsection{Model Overview}

The overall architecture of the proposed POG model is shown in Figure~\ref{fig:model_arch}. Three major components are included: 1) an autoencoder, 2) an adversarial generation module, and 3) an auxiliary classifier. 

The idea of generating effective pseudo OOD samples originates from this observation: most OOD samples look similar to IND inputs (i.e., sharing the same phrases or patterns) but do not correspond to any IND intents. Such OOD samples are usually harder to detect, and more efficient for improving the OOD detection performance if they are used to optimize the ER term \cite{lee2017training}. Therefore, in this study, we use the autoencoder to map an input utterance $x$ into a latent code $z$ and use the adversarial generation module to imitate the real latent code $z$ with a fake one $\tilde{z}$. In this way, the generated utterance $\tilde{x}'$ will look similar to the reconstructed utterance $x'$. Further, the auxiliary classifier is trained to predict the correct intent label associated with the reconstructed utterance $x'$ and used to regularize the latent code generator to make sure the generated utterance $\tilde{x}'$ to have indistinguishable intent labels and thus being OOD.

Each component is detailed below.

\subsubsection{Autoencoder}

This component intends to capture a continuous latent space for the input utterances so that we can apply gradient-based approaches more easily. It contains two functions: An encoder $\rm Enc_{\phi}$ (parameterized by ${\phi}$) that maps an input utterance $x$ to a latent code $z$, i.e.,
$$z={\rm Enc}_{\phi}(x),$$
and an decoder $P_{\psi}(x|z)$ (parameterized by ${\psi}$) that reconstructs an utterance $x'$ out of $z$. The autoencoder is trained by minimizing the reconstruction loss:
\begin{equation}\label{eq:reconst_loss}
  \begin{split}
    \mathcal{L}_{rec}(\phi, \psi) &= \mathop{\mathbb{E}}_{\epsilon \sim \mathcal{N}(0, \mathbf{I})}\mathop{\mathbb{E}}_{x \sim \mathcal{P}_{ind}}[-\log P_{\psi}(x|z + \epsilon)] \\
    &= \mathop{\mathbb{E}}_{\epsilon \sim \mathcal{N}(0, \mathbf{I})}\mathop{\mathbb{E}}_{x \sim \mathcal{P}_{ind}}[-\log P_{\psi}(x|{\rm Enc}_{\phi}(x) + \epsilon)]
  \end{split}
\end{equation}
Note that in order to smooth the latent space, we add a Gaussian noise $\epsilon \sim \mathcal{N}(0, \mathbf{I})$ to $z$ before feeding it to the decoder in the training phase, i.e., the input utterance $x$ is first encoded into a continuous vector $z$, and then a noise $\epsilon$ is sampled from the standard normal distribution $\mathcal{N}(0, \mathbf{I})$. A reconstructed utterance $x'$ is generated based on the latent code $z + \epsilon$ and the autoencoder is optimized to minimize the reconstruction loss shown in Eq.\ref{eq:reconst_loss}. This approach is reported to be effective for smoothing the produced latent space of an autoencoder~\cite{subramanian2018towards}.

\subsubsection{Adversarial Generation Module}

This component uses an adversarial training process to approximate the latent codes corresponding to the IND data. Two functions are involved: a generator $G_{\xi}$ which maps a noise $\epsilon \sim \mathcal{N}$ to a latent code $\tilde{z}$, i.e., 
$$\tilde{z} = G_{\xi}(\epsilon),$$ 
and a discriminator $D_{\eta}$ which distinguishes the \emph{real} latent code $z$ from the \emph{generated} latent code $\tilde{z}$. Intuitively, the generator aims to fool the discriminator, while the discriminator aims to discriminate real codes from generated ones. Once the training is done, the generated latent code $\tilde{z}$ is expected to be able to produce utterances that are similar to the IND inputs.

In this study, we train our generator and discriminator by minimizing the Wasserstein-1 distance~\cite{arjovsky2017wasserstein} between the generated distribution and the data distribution. Specifically, the loss for the generator is
\begin{equation}
    \mathcal{L}_{g}(\xi) = \mathop{\mathbb{E}}_{\epsilon \sim \mathcal{N}}[-D_{\eta}(G_{\xi}(\epsilon)]
\end{equation}
whereas the loss for the discriminator is:
\begin{equation}
    \mathcal{L}_{d}(\eta) = \mathop{\mathbb{E}}_{\epsilon \sim \mathcal{N}}[D_{\eta}(G_{\xi}(\epsilon)] - \mathop{\mathbb{E}}_{x \sim \mathcal{P}_{ind}}[D_{\eta}({\rm Enc}_{\phi}(x))]
\end{equation}
In order to enforce the 1-Lipschitz constraint~\cite{arjovsky2017wasserstein} on the discriminator $D_{\eta}$, we employ the gradient penalty term as proposed in \cite{gulrajani2017improved}:
\begin{equation}
\mathcal{L}_{gp}(\eta)=\mathop{\mathbb{E}}_{\hat{x} \sim \mathcal{P}_{\hat{x}}}[(\lVert \nabla_{\hat{x}}D_{\eta}(\hat{x}) \rVert_2 - 1)^2]
\end{equation}
where the sampling process from $\mathcal{P}_{\hat{x}}$ is approximated by uniformly interpolating between two random samples, one from the data distribution and another from the generated distribution~\cite{gulrajani2017improved}.

\subsubsection{Auxiliary Classifier}

This component of our model maps a decoded utterance to an $m$-dimension label distribution, where $m$ is the number of IND intent labels. The parameters $\omega$ of the \underline{a}uxiliary \underline{c}lassifier (AC) $P_{\omega}(\vec{y}|x')$ are optimized with the cross-entropy loss to predict the correct intent label associated with the utterances $x'$ decoded from the ``real'' latent code. Specifically, for an IND input $x_i$ (with intent label $y_i$) that is sampled from the training data, the corresponding ``real'' latent code $z_i={\rm Enc}_{\phi}(x_i)$ is first generated using the encoder and a reconstructed utterance $x_i'$ is decoded based on $z_i$. Then we feed the decoded utterance $x_i'$ to the AC and try to predict its intent label $y_i$. The following loss is optimized:
\begin{equation}
    \mathcal{L'}_{ce}(\omega) = \mathop{\mathbb{E}}_{
    \substack{
    (x_i,y_i) \sim \mathcal{P}_{ind}, \\ 
    x_i' \sim P_{\psi}(x|{\rm Enc}_{\phi}(x_i))}}
    [-log P_{\omega}(y=y_i|x_i')]
\end{equation}

Further, we use the AC to guide our latent code generator $G_{\xi}$ to produce latent codes that can be decoded into OOD samples, i.e., the generator $G_{\xi}$ is optimized to enforce a high entropy for the predicted distribution of the AC on the utterance $\tilde{x}'$ decoded from the ``fake'' latent code. Specifically, we first sample a noise $\epsilon \sim \mathcal{N}$ from the standard normal distribution and generate a ``fake'' code $\tilde{z}=G_{\xi}(\epsilon)$ using the generator. An utterance $\tilde{x}'$ is decoded based on $\tilde{z}$ (i.e., $\tilde{x}' \sim P_{\psi}(x|\tilde{z})$) and the following loss is optimized:
\begin{equation}\label{eq:reg_loss}
    \mathcal{L'}_{ent}(\xi) = \mathop{\mathbb{E}}_{
    \substack{
    \epsilon \sim \mathcal{N}, \\
    \tilde{x}' \sim P_{\psi}(x|G_{\xi}(\epsilon))}}
    [-\mathcal{H} (P_{\omega}(\vec{y}|\tilde{x}'))]
\end{equation}
where $\vec{y}$ denotes the intent type space. 

\begin{algorithm}[!t]
\caption{Training Procedure of POG}
\label{alg:train_pog}
\begin{algorithmic}[1]
\FOR{each training iteration}
\STATEx\hspace{\algorithmicindent} \textbf{/* (1) \emph{Train the autoencoder}} (Enc$_\phi$, Dec$_\psi$)
\STATEx\hspace{\algorithmicindent} \textbf{\emph{and the auxiliary classifier}} (AC$_\omega$) */
\STATE Sample ${\{x_i, y_i\}}_{i=1}^M \sim \mathcal{D}_{ind}$
\STATE Compute $z_i={\rm Enc}_{\phi}(x_i)$, $i=1, \cdots, M$
\STATE Add Gaussian noise $z_i \xleftarrow{} z_i + \epsilon$, $\epsilon \sim \mathcal{N}(0, \mathbf{I})$
\STATE Update $\phi, \psi$ by minimizing $\mathcal{L}_{rec}(\phi, \psi)$
\STATE Decode $x'_i \sim P_{\psi}(x|z_i)$\label{alg:step_dec_real}
\STATE Update $\omega$ by minimizing $\mathcal{L'}_{ce}(\omega)$

\STATEx\hspace{\algorithmicindent} \textbf{/* (2) \emph{Train the discriminator}} ($D_\eta$) */ 
\STATE Sample ${\{x_i, y_i\}}_{i=1}^M \sim \mathcal{D}_{ind}$, $\{\epsilon_i\}_{i=1}^M \sim \mathcal{N}$
\STATE Compute $z_i={\rm Enc}_{\phi}(x_i)$, and $\tilde{z}_i=G_{\xi}(\epsilon_i)$
\STATE Update $\eta$ by minimizing $\mathcal{L}_{d}(\eta) + \mathcal{L}_{gp}(\eta) $

\STATEx\hspace{\algorithmicindent} \textbf{/* (3) \emph{Train the generator}} ($G_\xi$) */
\STATE Sample $\{\epsilon_i\}_{i=1}^M \sim \mathcal{N}$
\STATE Compute $\tilde{z}_i=G_{\xi}(\epsilon_i)$
\STATE Update $\xi$ by minimizing $\mathcal{L}_{g}(\xi)$
\STATE Decode $\tilde{x}'_i \sim P_{\psi}(x|\tilde{z}_i)$\label{alg:step_dec_fake}
\STATE Update $\xi$ by minimizing $\mathcal{L'}_{ent}(\xi)$\label{alg:step_ce_loss}
\ENDFOR
\end{algorithmic}
\end{algorithm}

Note that the latent code generator $G_{\xi}$ trained using the loss $\mathcal{L}_{g}(\xi)$ and $\mathcal{L'}_{ent}(\xi)$ is trying to accomplish two adversarial targets: \textbf{First}, the adversarial loss $\mathcal{L}_{g}(\xi)$ forces the generated latent code to be close to the IND space, and thus makes the decoded utterance $\tilde{x}'$ looks similar to utterances in $\mathcal{D}_{ind}$; \textbf{Second}, the regularization loss $\mathcal{L'}_{ent}(\xi)$ ensures the intent associated with the decoded utterance $\tilde{x}'$ cannot be predicted by the AC. Specifically, $\mathcal{L'}_{ent}(\xi)$ reaches its minimum when the AC produces a uniform distribution, which means $\tilde{x}'$ does not belong to any existing intent labels. It is expected that these losses can guide our model to generate OOD samples near the IND distribution (i.e., look similar to the IND samples), thereby making the model more effective in OOD detection. 

The training process of the POG model is detailed in Algorithm~\ref{alg:train_pog}. Note that the text sample $\tilde{x}'_i$ which is decoded in Step~\ref{alg:step_dec_fake} of Algorithm~\ref{alg:train_pog} is discrete and non-differentiable, which hinders the gradients back-propagating from the AC to the generator $G_{\xi}$ through the loss $\mathcal{L'}_{ent}(\xi)$ in Step~\ref{alg:step_ce_loss} . In order to address this issue, we use a continuous approximation approach to replace the token (i.e., one-hot vector) sampled at each time step in $\tilde{x}'_i$ (in Step~\ref{alg:step_dec_fake}) and $x'_i$ (in Step~\ref{alg:step_dec_real}) with the probability vector produced by the decoder $P_{\psi}(x|z)$. These ``soft'' tokens are fed into the AC to make the whole computation process differentiable. Specifically, the word embedding fed at each time step of the AC is computed as an average over all the word embeddings weighted by the input probability distribution. 

Moreover, we also applied the temperature scaling technique to sharp the output distribution, i.e., the logits in each time step are divided by a temperature of $t$ to produce the output distribution. In our experiments, the value of $t$ decreases from 1 to 0 as the training proceeds. A small value of $t$ sharpens the output distribution to make it close to a one-hot vector.

\subsection{Utilizing Unlabeled Data}
It is observed in previous studies that if the OOD data used in the training and testing phase are similar, the OOD detection performance is better~\cite{hendrycks2018deep,lee2017training}. However, using human-labeled OOD data in Eq.~\ref{eq:entropy} is expensive, and it is hard for the POG model to gain any prior knowledge about the testing OOD samples if only IND data is utilized. However, fortunately, the unlabeled data $\mathcal{D}_{mix}$, which are usually easier to collect, can help us to solve this issue.

In this study, we augment the training process of the autoencoder in POG with the data from $\mathcal{D}_{mix}$. It is expected that the latent space produced by the autoencoder will be enriched with OOD samples, and the AC will lead the generator to focus on these OOD samples in the adversarial training process. Specifically, this involves adding a few additional steps in Algorithm~\ref{alg:train_pog}, in particular, two steps are added in each training iteration: (1b) update $\phi, \psi$ with the reconstruction loss and (2b) update $\eta$ with the discriminator loss utilizing the data sampled from $\mathcal{D}_{mix}$. We denote the augmented model as AEPOG, and the corresponding training algorithm is shown in Algorithm~\ref{alg:train_ae_pog}. Experiments in Section~\ref{sec:experiment} demonstrate that the generated pseudo OOD utterances can be more effective at improving the OOD detection performance. 

\begin{algorithm}[!t]
\caption{Training Procedure of AEPOG}
\label{alg:train_ae_pog}
\begin{algorithmic}[1]
\STATE Each training iteration additionally:
\STATEx \hspace{\algorithmicindent} \textbf{/* (1b) \emph{Train the autoencoder}} (Enc$_\phi$, Dec$_\psi$) */
\STATE \hspace{\algorithmicindent} Sample ${\{\hat{x}_i\}}_{i=1}^M \sim \mathcal{D}_{mix}$
\STATE \hspace{\algorithmicindent} Compute $z_i={\rm Enc}_{\phi}(\hat{x}_i)$, $i=1, \cdots, M$
\STATE \hspace{\algorithmicindent} Add Gaussian noise $z_i \xleftarrow{} z_i + \epsilon$, $\epsilon \sim \mathcal{N}(0, \mathbf{I})$
\STATE \hspace{\algorithmicindent} Update $\phi, \psi$ by minimizing $\mathcal{L}_{rec}(\phi, \psi)$

\STATEx\hspace{\algorithmicindent} \textbf{/* (2b) \emph{Train the discriminator}} ($D_\eta$) */
\STATE \hspace{\algorithmicindent} Sample ${\{\hat{x}_i\}}_{i=1}^M \sim \mathcal{D}_{mix}$, $\{\epsilon_i\}_{i=1}^M \sim \mathcal{N}$
\STATE \hspace{\algorithmicindent} Compute $z_i={\rm Enc}_{\phi}(\hat{x}_i)$, and $\tilde{z}_i=G_{\xi}(\epsilon_i)$
\STATE \hspace{\algorithmicindent} Update $\eta$ by minimizing $\mathcal{L}_{d}(\eta) + \mathcal{L}_{gp}(\eta)$
\end{algorithmic}
\end{algorithm}

%% file: sections/experiment.tex
\section{Experiment}\label{sec:experiment}

\subsection{Dataset}
We evaluated the proposed model on two datasets:
\begin{enumerate}
  \item \textbf{OSQ dataset}~\cite{emnlp19data}: this dataset covers 150 IND intents and also provides a set of manually labeled \underline{O}ut-of-\underline{S}cope \underline{Q}ueries (\textbf{OSQ}) that are not supported by the current system. These out-of-scope queries are regarded as OOD data in our experiments.
  \item \textbf{IPA dataset}: this dataset contains a set of Chinese utterances that were collected and annotated in the development process of a commercialized \underline{I}ntelligent \underline{P}ersonal \underline{A}ssistant (\textbf{IPA}) named Bixby\footnote{Bixby is shipped with a large number of Samsung products, such as phones, refrigerators, or watches.}. This dataset covers 67 domains, which can be further divided into 1,310 intents. In this study, 310 intents are randomly selected as the OOD intents, and the corresponding utterances are used as OOD data. Domain level labels are used for the IND utterances, i.e., 67 class labels are used for these IND data, and the utterances for each IND intent is assigned with the corresponding domain label.
\end{enumerate}

The IND data were divided into the train, validation, and test sets, while the OOD data were only used for validation and test. In addition to the labeled data, some unlabeled data (i.e., a mixture of IND and OOD data) $\mathcal{D}_{mix}$ were also collected for the AEPOG model. Specifically, for the OSQ dataset, a mixture of 10K IND data and 250 OOD data were used as $\mathcal{D}_{mix}$. For the IPA dataset, 20K unlabeled utterances were extracted from user logs and used as $\mathcal{D}_{mix}$. The partitions of the data used in our experiment are shown in Table~\ref{tab:dataset}.

Note that the data partition scheme for the IPA dataset to define OOD inputs ensures that the OOD and IND utterances have similar patterns. This simulates the actual situations that we have met in our real system, i.e., users may issue OOD utterances that are similar to IND utterances since they are not familiar with the precise capabilities of the system. This data partition scheme makes the OOD detection task more difficult, and it demonstrates a lower bound for the OOD detection performance of a real NLU system. Measuring and guaranteeing such lower bounds are essential for practical systems.

In order to better demonstrate the performance of our method on the IPA dataset, we collect an extra set of test data, i.e., Chat dataset, for the models trained on the IPA dataset. This dataset covers a broad range of OOD utterances such as chitchat or nonsense, which are commonly received by a practical NLU system. The Chat dataset contains 1.2K utterances, and it is only used in the test phase of our experiment. The utterances in the Chat dataset are manually annotated based on the user logs.

\begin{table}[!t]
\begin{center}
\caption{Statistics of the OSQ and IPA datasets. $\mathcal{D}_{mix}$ denotes a mixture of IND and OOD data.}\label{tab:dataset}
\begin{tabular}{lllll}
\toprule
 & & Train & Validate & Test\\
\midrule
 \multirow{3}{*}{\shortstack{OSQ\\Dataset}} & IND & 15.00K & 3.00K & 4.50K \\
 & OOD & - & 0.10K & 1.00K \\
 & $\mathcal{D}_{mix}$ & 10.25K & - & - \\
\midrule
 \multirow{3}{*}{\shortstack{IPA\\Dataset}} & IND & 28.90K & 3.60K & 3.60K \\
 & OOD & - & 1.20K & 1.20K \\
 & $\mathcal{D}_{mix}$ & 20.00K & - & - \\
\bottomrule
\end{tabular}
\end{center}
\end{table}

\begin{table}[!t]
\caption{OOD detection performance on the OSQ dataset. Each result in this table is an average of ten different runs. The notation $\uparrow$ means higher values are better, and $\downarrow$ means lower values are better. The model ER+AEPOG is significantly better than other model with $p$-value $<$ 0.01 ($\dag$) and $p$-value $<$ 0.05 ($*$) using t-test}\label{tab:res_without_ood_osq}
\begin{center}
\begin{tabular}{lllll}
\toprule
Model & AUROC$\uparrow$ & AUPR$\uparrow$ & FPR95$\downarrow$ & FPR90$\downarrow$ \\
\midrule
Cont. GAN        & $52.22^\dag$  & $82.79^\dag$  & $94.40^\dag$  & $88.17^\dag$  \\
Maha. Dis.       & $67.14^\dag$  & $90.56^\dag$  & $91.94^\dag$  & $83.30^\dag$  \\
Likelihood Ratio & $85.60^\dag$  & $96.07^\dag$  & $62.50^\dag$  & $43.40^\dag$  \\
AE               & $87.78^\dag$  & $96.98^\dag$  & $58.50^\dag$  & $40.10^\dag$  \\
MSP              & $92.86^\dag$  & $98.24^\dag$  & $39.72^\dag$  & $21.76^\dag$  \\
Entropy          & $92.82^\dag$  & $98.87^\dag$  & $31.91^\dag$  & $19.64^\dag$  \\
KNN              & $93.33^\dag$  & $98.20^\dag$  & $33.86^\dag$  & $18.78^\dag$  \\
DOC              & $94.24^\dag$  & $98.55^\dag$  & $30.02^\dag$  & $14.94^\dag$  \\
ODIN             & $95.14^*$     & $98.84^*$     & $26.04^*$     & $11.70^\dag$  \\
ER+Perturb       & $94.01^\dag$  & $98.55^\dag$  & $34.04^\dag$  & $15.32^\dag$  \\
ER+Mix           & $93.48^\dag$  & $98.31^\dag$  & $33.16^\dag$  & $17.86^\dag$  \\
\midrule
ER+POG           & $95.41^*$     & $98.94^*$     & $25.00^*$     & $10.10^\dag$  \\
ER+AEPOG         &$\textbf{95.83}$& $\textbf{99.05}$& $\textbf{23.70}$  & $\textbf{9.50}$  \\
\midrule
w.o. Noise       & $94.03^*$  & $98.53^*$  & $35.22^\dag$  &  $17.74^\dag$  \\
w.o. Soft Token  & $93.85^\dag$  & $98.53^*$  & $39.24^\dag$  &  $17.88^\dag$  \\
\bottomrule
\end{tabular}
\end{center}
\end{table}

\subsection{Model Implementation}\label{sec:model_imple}

Our text classifier was implemented using the CNN architecture~\cite{kim2014convolutional}. Four kernel sizes (2, 3, 4, 5) were used, and each kernel had 128 feature maps. A three-layer MLP was added on top of the CNN feature, and the hidden size of each layer was 512.  The vocabulary size was of 8.3K and 18.2K for the classifiers on the OSQ and IPA datasets, respectively. Note that the vocabulary we used covers all the tokens in the IND and OOD data. Pre-trained word vectors with the dimension of 100 were used to initialize the word embeddings. Specifically, 81.46\% and 70.01\% percent of word embeddings were initialized with the pre-trained word vectors for the IND and OOD datasets, respectively. The rest word embeddings were randomly initialized. All the classifier was trained for 30 epochs.

For the POG model, the auxiliary classifier shared the same structure and hyper-parameter setting with the text classifier. The encoder and decoder were both implemented using LSTM~\cite{hochreiter1997long}. The hidden size of the LSTM used for the OSQ and IPA datasets was 100 and 256, respectively. The generator and discriminator were both four-layer MLPs activated with the Leaky ReLU function~\cite{maas2013rectifier}. The MLP hidden size was 512 and 1024, and the vocabulary size was of 5.8K and 14.1K, respectively, for the OSQ and IPA datasets. It covers all the tokens in the IND training data. The POG and AEPOG models were trained for 80 epochs.

\subsection{Baselines}
In this work, we used several threshold-based OOD detectors as baselines. These baselines differ mainly in the way to calculate detection scores.

\begin{table}[!t]
\caption{OOD detection performance on the IPA dataset. Each result in this table is an average of ten different runs. The notation $\uparrow$ means higher values are better, and $\downarrow$ means lower values are better. The model ER+AEPOG is significantly better than other model with $p$-value $<$ 0.01 (marked by $\dag$) and $p$-value $<$ 0.05 ($*$) using t-test}\label{tab:res_without_ood_ipa}
\begin{center}
\begin{tabular}{lllll}
\toprule
Model & AUROC$\uparrow$ & AUPR$\uparrow$ & FPR95$\downarrow$ & FPR90$\downarrow$ \\
\midrule
Cont. GAN        & $58.06^\dag$  & $80.47^\dag$  & $91.51^\dag$  & $85.05^\dag$ \\
Maha. Dis.       & $45.65^\dag$  & $23.45^\dag$  & $95.01^\dag$  & $89.97^\dag$ \\
Likelihood Ratio & $69.65^\dag$  & $86.21^\dag$  & $84.44^\dag$  & $74.46^\dag$ \\
AE               & $67.57^\dag$  & $85.91^\dag$  & $86.77^\dag$  & $74.79^\dag$ \\
MSP              & $72.66^\dag$  & $86.42^\dag$  & $77.79^\dag$  & $86.42^\dag$ \\
Entropy              & $72.87^\dag$  & $86.42^\dag$  & $88.15^\dag$  & $75.41^\dag$ \\
KNN              & $67.94^\dag$  & $82.62^\dag$  & $76.87^\dag$  & $63.43^\dag$ \\
DOC              & $71.68^\dag$  & $46.03^\dag$  & $79.40^\dag$  & $64.34^\dag$ \\
ODIN             & $72.90^\dag$  & $86.53^\dag$  & $77.57^\dag$  & $63.28^\dag$ \\
ER+Perturb       & $73.24^\dag$  & $86.89^*$  & $77.88^\dag$  & $63.30^\dag$ \\
ER+Mix           & $33.71^\dag$  & $63.55^\dag$  & $96.44^\dag$  & $92.33^\dag$ \\
\midrule
ER+POG           & $73.83^\dag$  & $87.15^*$  & $76.17^\dag$  & $62.28^\dag$ \\
ER+AEPOG         & $\textbf{75.86}$  & $\textbf{87.95}$  & $\textbf{71.67}$  & $\textbf{56.78}$ \\
\midrule
w/o Noise        & $71.01^*$  & $85.10^*$  & $79.17^\dag$  &  $65.17^\dag$  \\
w/o Soft Token   & $72.09^*$  & $86.31^*$  & $80.22^\dag$  &  $65.07^\dag$  \\
\bottomrule
\end{tabular}
\end{center}
\end{table}

\begin{enumerate}
  \item \textbf{MSP}~\cite{hendrycks2016baseline}: A text classifier with a Softmax output layer is trained, and the maximum Softmax output is used as the detection score (i.e., calculated using Eq.~\ref{eq:ood_score}). The training objective of this classifier does not include the entropy regularization (ER) term.

  \item \textbf{Entropy}: The Shannon entropy of the predicted distribution for each input is used as the detection score. Higher entropy means higher uncertainty of the prediction, which indicates that the input sample may be from the OOD data.
  
  \item \textbf{ODIN}~\cite{liang2017enhancing}: The temperature scaling and input perturbation technique is applied to the text classifier as obtained in the previous \textbf{MSP} baseline, and the maximum Softmax output is used as the detection score. Note that small perturbations of each input are added to the last feature layer in this baseline.

  \item \textbf{DOC}~\cite{shu2017doc}: $m$ binary classifiers are built for $m$ classes. The maximum confidence score predicted by these $m$ classifiers is used as the detection score.

  \item \textbf{KNN}: The feature for each input sample is extracted using a pre-trained classifier (i.e., the classifier as obtained in the previous \textbf{MSP} baseline), and the Euclid distance of each feature vector to its nearest class is used as the detection score.
  
  \item \textbf{Maha. Dis.}~\cite{Lee2018A}: The Mahalanobis distance is used to replace the Euclid distance in the previous \textbf{KNN} baseline.
  
  \item \textbf{Cont. GAN}~\cite{ryu2018out}: A GAN-based model is trained to generate continuous OOD features to mimic features extracted from IND samples. The confidence of the discriminator is used as the detection score. A low confidence score suggests that the input sample may be from the OOD data.
  
  \item \textbf{AE}: An autoencoder is trained on the IND data, and the reconstruction error of each input is used as the detection score.
  
  \item \textbf{Likelihood Ratio}~\cite{ren2019likelihood}: The likelihood ratio is used as the detection score. The background model is trained using perturbed IND samples, which are generated by randomly replacing tokens in the IND samples with a probability of 0.5\footnote{Other probability values were also tested, but similar results were obtained}.  
  
  \item \textbf{ER+Perturb}: An ER term $\mathcal{L}_{ent}(\theta)$ (Eq.~\ref{eq:entropy}) is added to the training objective of the classifier, and this term is optimized using perturbed IND samples. These perturbed samples are constructed in a similar way as in the previous \textbf{Likelihood Ratio} baseline.
  
  \item \textbf{ER+Mix}: The ER term in the training objective is optimized using the data from $\mathcal{D}_{mix}$. This baseline demonstrates a naive way to utilize $\mathcal{D}_{mix}$.
\end{enumerate}

Our baselines cover a variety of competitive OOD detection models that are currently available.  Note that the text classifier involved in each baseline follows the same structure and hyper-parameter setting with our classification model (see Section~\ref{sec:model_imple}), and the baseline \textbf{AE} and \textbf{Likelihood Ratio} share the same LSTM structure with our POG model.

Several ablation tests are also performed to validate the effect of each component in our model:
\begin{enumerate}
    \item \textbf{w/o Noise}: The autoencoder is trained without adding the Gaussian noise in Eq.~\ref{eq:reconst_loss}.
    \item \textbf{w/o Soft Token}: The loss $\mathcal{L'}_{ent}(\xi)$ in Eq.~\ref{eq:reg_loss} is optimized without using the soft token approximation approach. We instead use the policy gradient algorithm to estimate the gradients through the sampling process of the decoder.
\end{enumerate}
Note that all the ablation studies are performed without using the unlabeled data $\mathcal{D}_{mix}$.

\begin{table}[!t]
\caption{OOD detection performance on the IPA dataset when testing with the Chat dataset. Each result in this table is an average of ten different runs. The notation $\uparrow$ means higher values are better, and $\downarrow$ means lower values are better. The model ER+AEPOG is significantly better than other model with $p$-value $<$ 0.01 (marked by $\dag$) and $p$-value $<$ 0.05 ($*$) using t-test}\label{tab:res_without_ood_ipa_chatbotdata}
\begin{center}
\begin{tabular}{lllll}
\toprule
Model & AUROC$\uparrow$ & AUPR$\uparrow$ & FPR95$\downarrow$ & FPR90$\downarrow$ \\
\midrule
Cont. GAN        & $69.80^\dag$  & $87.55^\dag$  & $90.77^\dag$  & $76.18^\dag$ \\
Maha. Dis.       & $73.85^\dag$  & $48.20^\dag$  & $73.96^\dag$  & $61.48^\dag$ \\
Likelihood Ratio & $90.83^\dag$  & $96.63^\dag$  & $41.51^\dag$  & $27.37^\dag$ \\
AE               & $92.41^\dag$  & $97.21^\dag$  & $37.77^\dag$  & $18.97^\dag$ \\
MSP              & $88.41^\dag$  & $95.05^\dag$  & $52.30^\dag$  & $32.56^\dag$ \\
Entropy          & $89.02^\dag$  & $95.05^\dag$  & $37.56^\dag$  & $29.31^\dag$ \\
KNN              & $89.43^\dag$  & $95.02^\dag$  & $36.17^\dag$  & $22.91^\dag$ \\
DOC              & $93.03^\dag$  & $97.28^\dag$  & $35.44^\dag$  & $18.07^\dag$ \\
ODIN             & $90.25^\dag$  & $95.67^\dag$  & $37.92^\dag$  & $24.29^\dag$ \\
ER+Perturb       & $96.28^\dag$  & $98.72^*$     & $19.23^\dag$  & $10.80^\dag$ \\
ER+Mix           & $92.76^\dag$  & $97.27^\dag$  & $38.44^\dag$  & $21.71^\dag$ \\
\midrule
ER+POG           & $98.53^*$     & $99.42^*$     & $7.70^*$      & $4.36^*$ \\
ER+AEPOG         & $\textbf{98.60}$  & $\textbf{99.46}$  & $\textbf{7.64}$  & $\textbf{4.24}$ \\
\midrule
w/o Noise        & $94.26^\dag$  & $98.07^\dag$  & $33.08^\dag$  &  $18.56^\dag$  \\
w/o Soft Token   & $94.08^\dag$  & $98.06^\dag$  & $40.07^\dag$  &  $16.96^\dag$  \\
\bottomrule
\end{tabular}
\end{center}
\end{table}

\subsection{Metrics}
For OOD detection, we used three common metrics~\cite{hendrycks2016baseline,hendrycks2018deep,ren2019likelihood,lee2017training}:

\begin{enumerate}
  \item \textbf{AUROC}. The area under the \underline{r}eceiver \underline{o}perating \underline{c}haracteristic (ROC) curve; Higher values are better.

  \item \textbf{AUPR}. The area under the precision-recall curve when OOD inputs are treated as the positive samples; Higher values are better.

  \item \textbf{FPR$N$}. The false-positive rate (FPR) when the true positive rate (TPR) is $N\%$\footnote{Note that FPR=$\frac{\rm FP}{\rm FP+TN}$, and TPR=$\frac{\rm TP}{\rm TP+FN}$, where FP, TP, TN, FN is the number of false positives, true positives, true negatives, false negatives, respectively.}. In this study, FPR95 and FPR90 are reported, since such settings are commonly used in practical systems. Lower values are better.
\end{enumerate}

Note that the metric FPR$N$ is of more practical value in real-world applications since it evaluates the performance of an OOD detection module at a particular threshold. It is directly related to the performance of a deployed system. Lower FPR$N$ means triggering fewer false alarms when the performance on the IND data is guaranteed with a precision of $N$\%~\cite{hendrycks2018deep}. On the contrary, the metrics AUPR and AUROC evaluate the performance across various thresholds. In other words, AUPR and AUROC are threshold independent.

\begin{figure}[!t]
  \centering
  \begin{tabular}{c}
  \leftskip 4em
    \includegraphics[width=6.3cm]{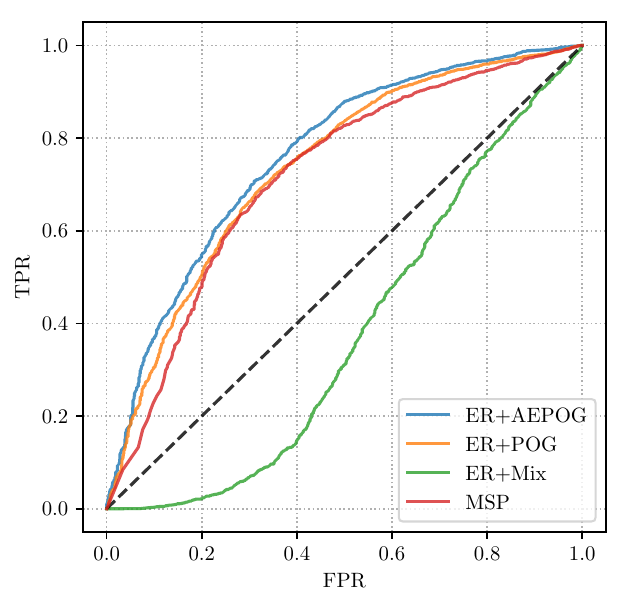}
  \end{tabular}
  \caption{ROC curves for the models on the IPA dataset. The model ER+AEPOG improves the AUROC of OOD detection compared to ER+POG, whereas a sharp decrease for the AUROC of OOD detection is observed for the model ER+Mix, which directly utilize $\mathcal{D}_{mix}$ to optimize the ER term of the classifier.}\label{fig:bixby_roc}
\end{figure}

\begin{figure}[!t]
  \centering
  \begin{tabular}{c}
  \leftskip 4em
    \includegraphics[width=6.4cm]{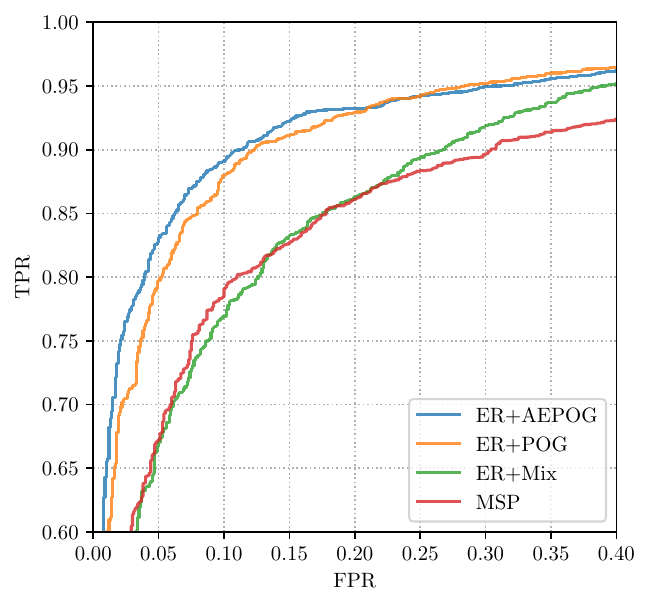}
  \end{tabular}
  \caption{ROC curves for the models on the OSQ dataset. We zooms in the upper-left corner of the ROC curves to facilitate a clearer view.}\label{fig:osq_roc}
\end{figure}

\subsection{Effects of Generated Pseudo OOD Utterances}
First of all, we evaluated the effectiveness of the pseudo OOD utterances generated by our model POG when only IND data is available. More concretely, the POG model was trained and, meanwhile, a set of pseudo OOD utterances (with the same size of the IND training data) was sampled to optimize the ER term $\mathcal{L}_{ent}(\theta)$ (i.e., Eq.~\ref{eq:entropy}). The performance of the resulting model (i.e., ER+POG) on the OSQ and IPA datasets is depicted in Table~\ref{tab:res_without_ood_osq} and Table~\ref{tab:res_without_ood_ipa}, respectively. Table~\ref{tab:res_without_ood_ipa_chatbotdata} shows the test results on the Chat dataset for the models trained on the IPA dataset.

It can be seen that our model outperforms all the baselines on both datasets significantly, which demonstrates the effectiveness of the generated pseudo OOD utterances. It is also interesting to see that:
\textbf{1)} Detecting OOD samples on the IPA dataset is more difficult compared to the OSQ dataset (see Table~\ref{tab:res_without_ood_osq} and Table~\ref{tab:res_without_ood_ipa}). It is because that most OOD inputs in the IPA test data look similar to the IND inputs. Moreover, the models trained on the IPA dataset obtain a large performance gain when testing on the Chat dataset, especially on the FPR$N$ metric. This validates our claim that the OOD utterances that look similar to IND utterances are more difficult to detect. Further, the fact that the proposed model ER+POG surpasses all the baselines on both datasets also proves its ability to handle various kinds of inputs with different distributions.
\textbf{2)} The results for the ablation studies validate the effect of the Gaussian noise and soft token approximation approach used in our study. A significant performance drop is observed on both datasets.
\textbf{3)} The proposed ER+POG model obtains large improvements on the FPR$N$ metrics. In particular, the relative performance gain for FPR95 and FPR90 on the OSQ dataset reaches 4.16\% and 15.84\%, respectively, compared to the best performing baselines. Even on the more difficult IPA dataset, the relative performance gain also reaches 0.92\% and 1.61\% for FPR95 and FPR90, respectively. This indicates that the proposed model is more suitable in real-world applications since the metric FPR$N$ directly reflects the performance of the deployed models.
\textbf{4)} The Cont. GAN baseline proposed in \cite{ryu2018out} only performs slightly better than random guesses. This shows that the confidence score of the discriminator is not efficient in detecting OOD inputs, and using GAN to generate continuous features is not helpful to improve the OOD detection performance in this baseline.

It is also worth mentioning that the performance of the NLU model on the IND input classification task is not affected if we optimize the ER term with utterances generated by the proposed model (see Table~\ref{tab:ind_acc}). Notably, in some cases, the utterances generated by the proposed model even help to improve the classification accuracy of the NLU module on IND inputs. This may be because optimizing the ER term with utterances produced by the POG or AEPOG model helps to prevent the classifier from over-fitting, and it further demonstrates the effectiveness of the proposed model in practical applications.

\begin{table}[!t]
\caption{Classification accuracy on IND inputs for each model. Each result in this table is an average of ten different runs. The model ER+AEPOG is significantly better than most of other models with $p$-value $<$ 0.01 (marked by $\dag$) and $p$-value $<$ 0.05 ($*$) using t-test}\label{tab:ind_acc}
\begin{center}
\begin{tabular}{lcc}
\toprule
Model & OSQ Dataset & IPA Dataset \\
\midrule
Cont. GAN        & $70.43^\dag$  & $84.68^\dag$   \\
Maha. Dis.       & $82.36^\dag$  & $89.73^\dag$   \\
MSP              & $92.61^*$     & $91.19$   \\
KNN              & $81.05^\dag$  & $88.32^\dag$   \\
DOC              & $91.84^\dag$  & $89.20^\dag$   \\
ER+Perturb       & $90.96^\dag$  & $91.22$   \\
ER+Mix           & $89.77^\dag$  & $90.41\dag$   \\
ER+POG           & $93.31$       & $91.37$  \\
ER+AEPOG         & \textbf{93.32}& \textbf{91.40}  \\
\midrule
w/o Noise        & $93.01$       & $91.34$ \\
w/o Soft Token   & $93.10$       & $90.96^*$ \\
\bottomrule
\end{tabular}
\end{center}
\end{table}

\begin{figure}[!t]
  \centering
  \begin{tabular}{c}
    \includegraphics[width=8.6cm]{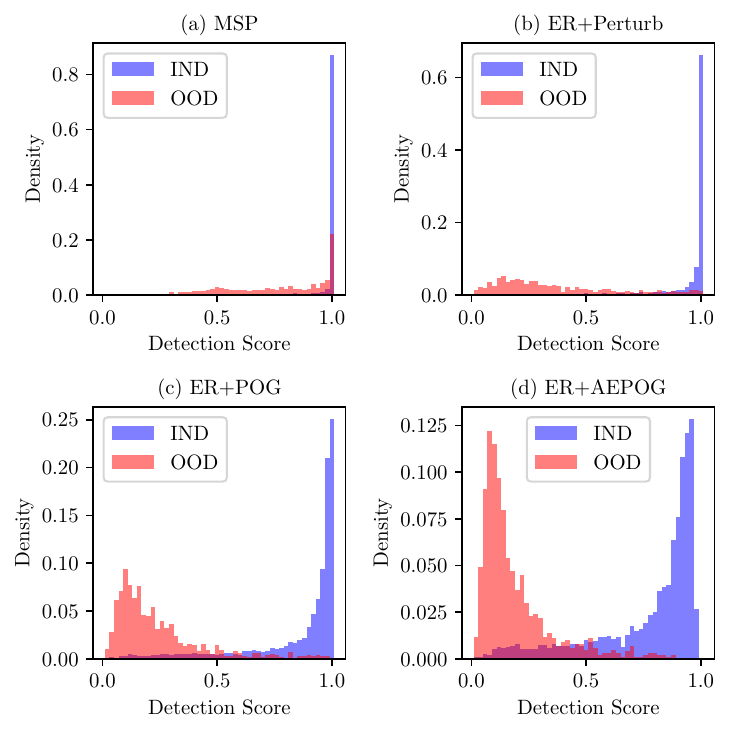} 
  \end{tabular}
  \caption{Distributions of detection scores corresponding to the IND and OOD samples of the OSQ dataset.}
  \label{fig:vis_distribution_OSQ}
\end{figure}

\begin{figure}[!t]
  \centering
  \begin{tabular}{c}
    \includegraphics[width=8.6cm]{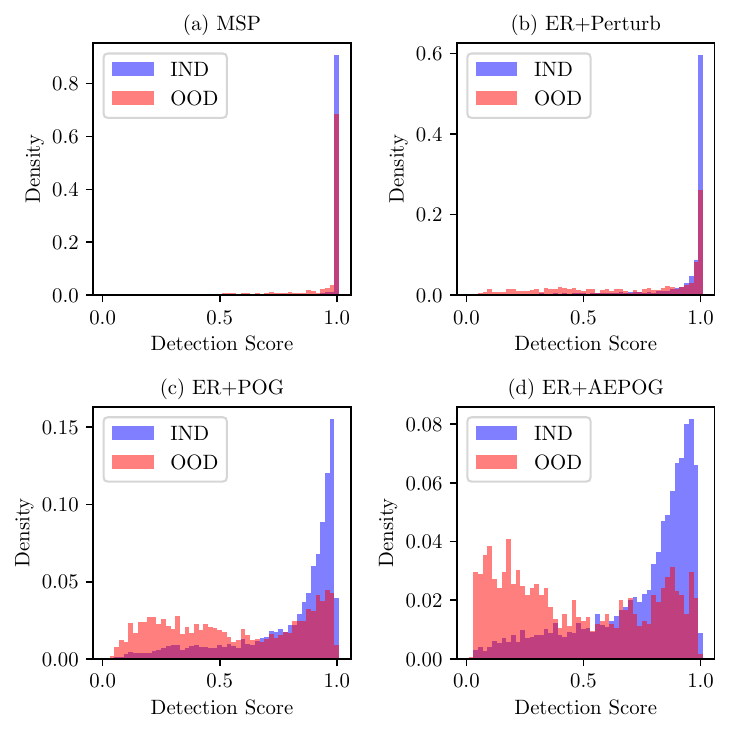} 
  \end{tabular}
  \caption{Distributions of detection scores corresponding to the IND and OOD samples of the IPA dataset.}
  \label{fig:vis_distribution_Bixby}
\end{figure}

\subsection{Effects of Utilizing Unlabeled Data}

We also verified the effectiveness of our model when a set of unlabeled data $\mathcal{D}_{mix}$ is available. Specifically, an AEPOG model was trained by optimizing $\mathcal{L}_{rec}(\phi, \psi)$ with the data from $\mathcal{D}_{mix}$ (Algorithm~\ref{alg:train_ae_pog}), and the sampled pseudo OOD utterances were used to optimize $\mathcal{L}_{ent}(\theta)$ when training the intent classifier. The performance of the resulting classifier on the OSQ and IPA datasets is shown in the last line of Table~\ref{tab:res_without_ood_osq} and Table~\ref{tab:res_without_ood_ipa}, respectively.

It can be seen that the model ER+AEPOG surpasses all the other models significantly, especially on the FPR$N$ metric. Particularly, on the OSQ dataset, the performance of the ER+AEPOG model improves 9.87\% and 23.16\% (relatively) compared to the best performing baselines on FPR95 and FPR90, respectively. On the IPA dataset, the performance improves 7.26\% and 11.45\% (relatively) for FPR95 and FPR90, respectively. This verifies our claim that the proposed AEPOG model can effectively utilize unlabeled data to improve the effectiveness of the generated OOD samples. Thus it is more suitable in practical applications.

It is also interesting to see that the baseline model ER+Mix, which optimizes the ER term $\mathcal{L}_{ent}(\theta)$ directly using the data from $\mathcal{D}_{mix}$ when training the classifier, experiences a remarkable performance drop on all metrics, especially on the IPA dataset (see Figure~\ref{fig:bixby_roc}). This indicates that the ER term itself cannot utilize $\mathcal{D}_{mix}$ directly. It further demonstrates that the performance improvement obtained by the ER+AEPOG model attributes to the proposed AEPOG model since the proposed adversarial training process helps to produce more effective pseudo OOD samples to improve the OOD detection performance.

Figure~\ref{fig:osq_roc} shows the ROC curves obtained from these models on the OSQ dataset. The proposed ER+POG model surpasses all the baselines, and ER+AEPOG further improves the performance of OOD detection. Similar results are also obtained on the IPA dataset (see Figure~\ref{fig:bixby_roc}). Note that the performance improvement of the ER+AEPOG model over the ER+POG model is relatively small on the OSQ dataset compared to the IPA dataset. This is partly because the OOD data only takes a small proportion (i.e., 2.44\%) of $\mathcal{D}_{mix}$ in the OSQ dataset, whereas about one third (33.33\%) intents are randomly selected as OOD in the IPA dataset.

We also analyzed the distribution of the detection scores obtained from different models on the OSQ and IPA datasets. Specifically, for the MSP model, the distributions corresponding to IND and OOD inputs are closely overlapped (see Figure~\ref{fig:vis_distribution_OSQ}a and Figure~\ref{fig:vis_distribution_Bixby}a). Marginal improvements are observed for the model ER+Perturb, which optimizes the ER term using randomly perturbed IND inputs (Figure~\ref{fig:vis_distribution_OSQ}b and Figure~\ref{fig:vis_distribution_Bixby}b). This indicates that the ER term helps to improve the OOD detection performance, but only to a limited extent. Better separation of the detection scores between the IND and OOD inputs is observed when the ER term is optimized with the pseudo OOD utterances generated by the POG model (Figure~\ref{fig:vis_distribution_OSQ}c and Figure~\ref{fig:vis_distribution_Bixby}c), and this separation is enlarged if the utterances generated by the AEPOG model are used (Figure~\ref{fig:vis_distribution_OSQ}d and Figure~\ref{fig:vis_distribution_Bixby}d). This indicates that the pseudo OOD utterances generated by the proposed model facilitate OOD detection, and the effectiveness of these pseudo OOD utterances can be further improved when utilizing the unlabeled data.

\begin{figure}[!t]
  \centering
  \begin{tabular}{c}
    \includegraphics[width=8.2cm]{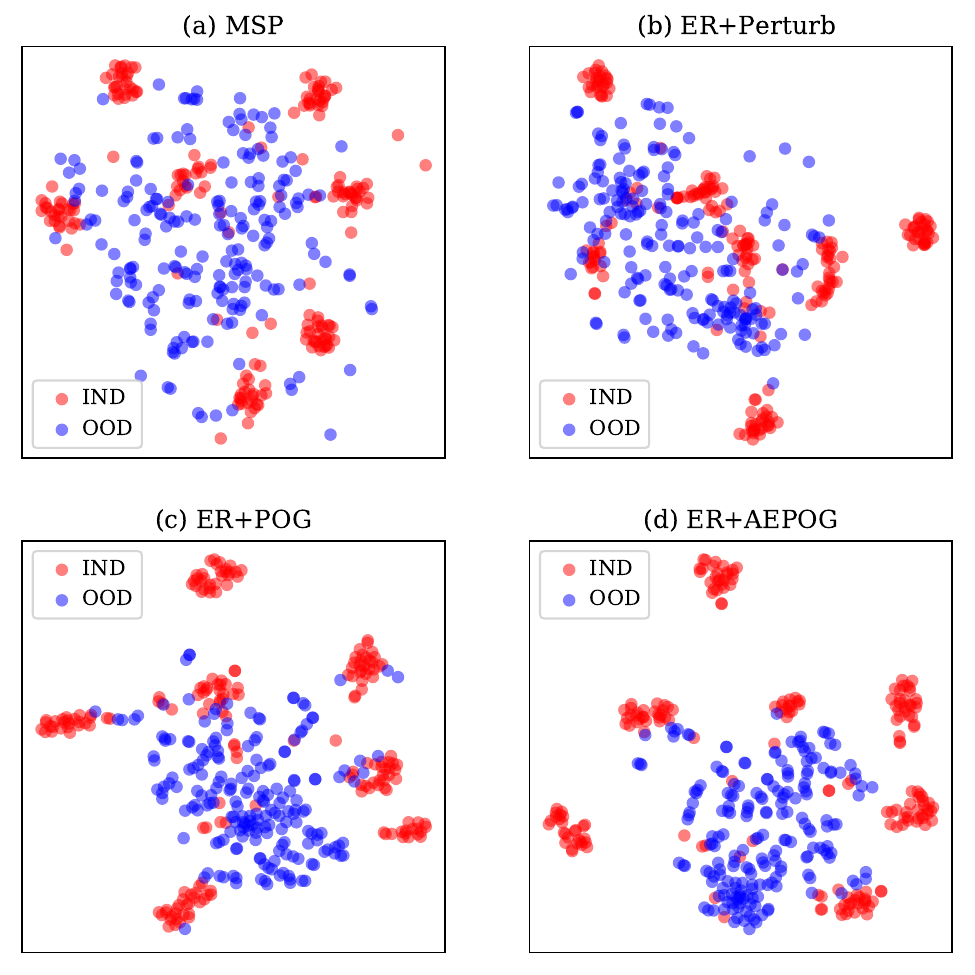} 
  \end{tabular}
  \caption{t-SNE visualization of the features vectors associated with the test samples from the OSQ dataset.}
  \label{fig:vis_latent}
\end{figure}

\subsection{Feature Vectors for IND and OOD Inputs}

To further investigate the benefit of the proposed model on OOD detection, we visualized the feature spaces of different intent classifiers on the OSQ dataset (see Figure~\ref{fig:vis_latent}). Specifically, we fed the test samples of the OSQ dataset to each intent classifier and obtained the feature vector of each sample from the penultimate layer of the network. The t-SNE algorithm~\cite{maaten2008visualizing} was used to map these vectors into 2-dimensions. 

It can be seen that all the intent classifiers cluster features of the IND samples into separated groups. This coincides with the observation that all models obtain high classification performance on IND inputs (Table~\ref{tab:ind_acc}). However, as to the OOD samples, the MSP model scatters the features of OOD samples around the features of IND samples (Figure~\ref{fig:vis_latent}a), making it hard for OOD detection. Whereas, the classifiers that are regularized by the ER term produce more distinguishable features for IND and OOD samples (Figure~\ref{fig:vis_latent}b). The utterances generated by the proposed POG and AEPOG model enhance this separation (Figure~\ref{fig:vis_latent}c and Figure~\ref{fig:vis_latent}d). This facilitates the detection of OOD inputs.

\subsection{Case Study for Generated OOD Utterances}

Table~\ref{tab:POG_sample} shows some randomly selected cases of the pseudo OOD utterances generated by the proposed POG and AEPOG model on the OSQ dataset. It is interesting to see that the POG model can combine words and phrases from the IND samples (such as ``French'' or ``my phone'') in a grammatical way. This makes the produced pseudo OOD utterances look similar to the IND data but possess indistinguishable intents. Moreover, the AEPOG model can make use of phrases from the OOD data, and thus produces more effective utterances that can be used to improve the OOD detection performance.

\begin{table}[!t]
\begin{center}
\caption{Utterances sampled from the IND and OOD test set of the OSQ dataset (i.e., human generated utterances), and the pseudo OOD utterances generated using the POG and AEPOG model.}\label{tab:POG_sample} 
\begin{tabular}{cl}
\toprule
\multirow{5}*{IND Samples} 
                          & Translate hello in French. \\
                          & Locate my phone please. \\
                          & Schedule a gas bill payment. \\
                          & Help me change my insurance plan. \\
                          & I'd like to improve my credit score. \\
\midrule
\multirow{5}*{OOD Samples} 
                          & How much is my car worth used. \\
                          & Can you add a bag to my reservation. \\
                          & How do you fix a leaking sink. \\
                          & How long do wire transfer take. \\
                          & When was Toyota created. \\
\midrule
\multirow{5}*{Generated by POG} & Please obtain French. \\
                          & How can make my phone get. \\
                          & What meetings can I schedule there. \\
                          & Can you help me an setting. \\
                          & I'd like to include the email my my dinner. \\
                          
\midrule
\multirow{5}*{Generated by AEPOG} & How much is this payments please. \\
                             & How good is my reservation like. \\
                             & How do you divided for pork. \\
                             & Tell me how long I taken for chilis off. \\
                             & When was today's name with delta vehicle. \\
\bottomrule
\hline
\end{tabular}
\end{center}
\end{table}

%% file: sections/conclusion.tex
\section{Conclusion}

In this paper, we propose a novel model POG to generate pseudo OOD samples that can be used to improve the performance of OOD detection in an NLU module. An autoencoder is used to map each input utterance to a latent code, and an adversarial training process is employed to make the codes of OOD samples similar to those of IND samples. This makes the generated OOD samples look similar to IND samples. Further, an auxiliary classifier is introduced to regularize the generator and thereby ensures these generated pseudo OOD samples to have indistinguishable intents. Our experiments show that the pseudo OOD samples generated by the proposed POG model can be used to effectively improve the OOD detection performance of the NLU module by optimizing the ER term when training the NLU module. Further, it is also demonstrated that augmenting the training process of the autoencoder in the POG model improves the effectiveness of the generated pseudo OOD samples.

As future works, we will explore this idea in classification problems for longer text since it is more challenging to generate longer documents for OOD detection. Also, note that our study focuses on detecting OOD inputs for the NLU module. It is worth exploring to equip other components of the task-oriented dialogue system, such as dialogue state tracking or dialogue management modules, with the ability to detect OOD inputs.